\documentclass[11pt,a4paper]{article}
\usepackage[hyperref]{eacl2021}
\usepackage{times}
\usepackage{latexsym,amsmath,amssymb}

\usepackage{booktabs,graphicx,multirow}
\usepackage{subcaption}
\usepackage{mathtools}
\DeclarePairedDelimiter\abs{\lvert}{\rvert}

\def\ie{i.e.\ }

\newcommand\ra{$\rightarrow$}
\newcommand{\MC}[3]{\multicolumn{#1}{#2}{#3}}

\newcommand{\B}{\textbf}
\newcommand{\I}{\textit}
\newcommand{\T}{\texttt}

\definecolor{drop}{HTML}{933D4E}
\definecolor{gain}{HTML}{32935D}
\definecolor{tabtitle}{HTML}{064791}
\newcommand{\ua}[1]{\textcolor{gain}{\small$\Uparrow {\B{#1}}$}}
\newcommand{\da}[1]{\textcolor{drop}{\small$\Downarrow {\B{#1}}$}}

\newcommand{\sta}[2]{\footnotesize{#1 $\pm$ #2}\normalsize}
\newcommand{\mask}[0]{\T{[MASK]\,}}

\usepackage{microtype}

\aclfinalcopy 
\def\aclpaperid{497} 

\title{Cross-lingual Visual Pre-training for Multimodal Machine Translation}

\author{
    Ozan Caglayan$^{1}${\normalfont,}
    Menekse Kuyu$^{2}${\normalfont,}
    Mustafa Sercan Amac$^{2}${\normalfont,}
    Pranava Madhyastha$^1$ \\[.2em]
    {\bf Erkut Erdem$^2${\normalfont,}
    Aykut Erdem$^3$ \and Lucia Specia$^{1,4,5}$}\\[.3em]
    Imperial College London$^1$,\,
    Hacettepe University$^2$,\,
    Koç University$^3$ \\[.2em]
    University of Sheffield$^4$,\,
    ADAPT - Dublin City University$^5$\\
    \texttt{\small o.caglayan@ic.ac.uk, meneksekuyu@gmail.com, sercanamac@gmail.com, pranava@ic.ac.uk} \\ \texttt{\small erkut@cs.hacettepe.edu.tr, aerdem@ku.edu.tr, l.specia@ic.ac.uk}\\
}

\date{}

\begin{document}
\maketitle

\begin{abstract}
Pre-trained language models have been shown to improve performance in many natural language tasks substantially. Although the early focus of such models was single language pre-training, recent advances have resulted in cross-lingual and visual pre-training methods. In this paper, we combine these two approaches to learn visually-grounded cross-lingual representations. Specifically, we extend the translation language modelling~\cite{xlm} with masked region classification and perform pre-training with three-way parallel vision \& language corpora. We show that when fine-tuned for multimodal machine translation, these models obtain state-of-the-art performance. We also provide qualitative insights into the usefulness of the learned grounded representations.
\end{abstract}

\section{Introduction}
\label{sec:intro}
Pre-trained language models~\cite{peters-etal-2018-deep,devlin-etal-2019-bert} have  been proven valuable tools for contextual representation extraction. Many studies have shown their effectiveness in discovering linguistic structures~\cite{tenney-etal-2019-bert}, which is useful for a wide variety of NLP tasks~\cite{talmor-etal-2019-commonsenseqa,kondratyuk-straka-2019-75,petroni2019language}. These positive results led to further exploration of (i)
cross-lingual pre-training~\cite{xlm,conneau-etal-2020-unsupervised,wang2020minilm} through the use of multiple mono-lingual and parallel resources, and (ii) visual pre-training where large-scale image captioning corpora are used to induce grounded vision \& language representations~\cite{vilbert,tan-bansal-2019-lxmert,unicodervl,vlbert,oscar}. The latter is usually achieved by extending the masked language modelling (MLM) objective~\cite{devlin-etal-2019-bert} with auxiliary vision \& language tasks such as masked region classification and image sentence matching.

In this paper, we present the first attempt to bring together cross-lingual and visual pre-training. Our visual translation language modelling (VTLM) objective combines the translation language modelling (TLM)~\cite{xlm} with masked region classification (MRC)~\cite{uniter,vlbert} to learn grounded cross-lingual representations. Unlike most of the prior work that use classification or retrieval based downstream evaluation, we focus on the \I{generative} task of multimodal machine translation (MMT), where images accompany captions during translation~\cite{sulubacak-etal-2020-multimodal}.
Once pre-trained, we transfer the VTLM encoder to a Transformer-based ~\cite{transformers} MMT and fine-tune it for the MMT task. To our knowledge, this is also the first attempt of pre-training \& fine-tuning for MMT, where the current state of the art mostly relies on training multimodal sequence-to-sequence systems from scratch~\cite{calixto-etal-2016-dcu,caglayan-etal-2016-multimodality,libovicky-helcl-2017-attention,elliott-kadar-2017-imagination,caglayan-etal-2017-lium,graph_based_att}.

Our findings highlight the effectiveness of cross-lingual visual pre-training: when fine-tuned on the English\ra German direction of the Multi30k dataset~\cite{elliott-etal-2016-multi30k}, our MMT model surpasses our constrained MMT baseline by about $10$ BLEU and $8$ METEOR points.
The rest of the paper is organised as follows: $\S$\ref{sec:method} describes our pre-training and fine-tuning protocol, $\S$\ref{sec:results} presents our quantitative and qualitative analyses, and $\S$\ref{sec:conclusion} concludes the paper with pointers for future work.

\begin{figure*}[t]
\centering
\includegraphics[width=0.99\textwidth]{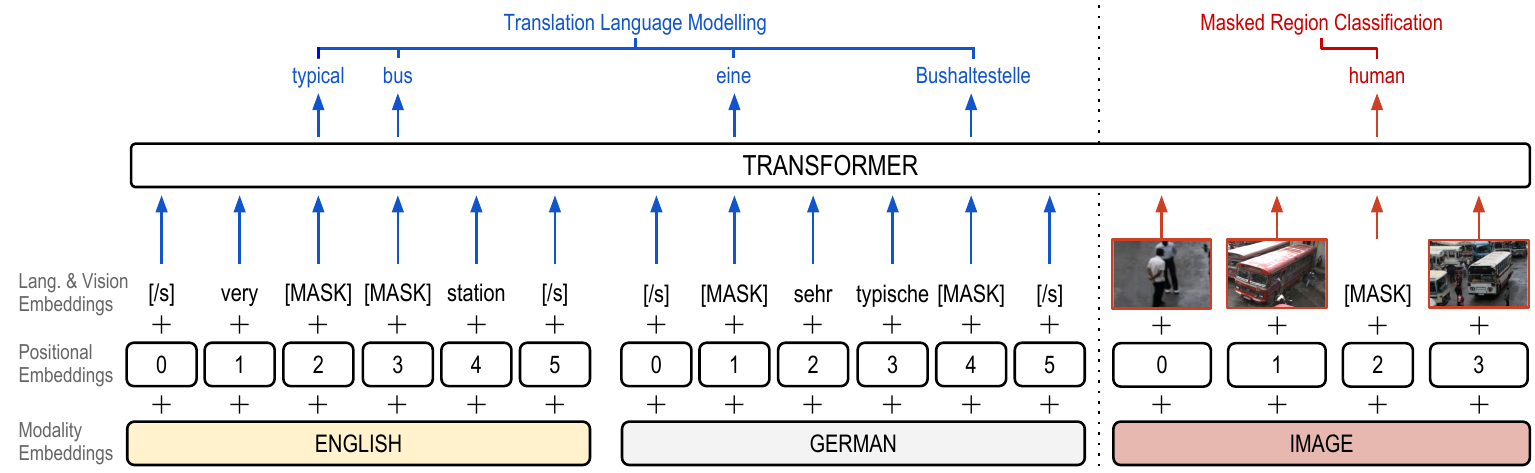}
\caption{The architecture of the proposed model: VTLM extends the TLM~\cite{xlm} (left side of the dotted line) with regional image features. Masking applies on both linguistic and visual tokens.}
\label{fig:VTLM}
\end{figure*}

\section{Method}
\label{sec:method}
We propose Visual Translation Language Modelling (VTLM) objective to learn multimodal cross-lingual representations. In what follows, we first describe the TLM objective~\cite{xlm} and then introduce the modifications required to extend it to VTLM.

\subsection{Translation language modelling}
\label{sec:tlm}
The TLM objective is based on Transformer networks and assumes the availability of parallel corpora during training. It defines the input $x$ as the concatenation of $m$-length source language sentence $s^{(1)}_{1:m}$ and $n$-length target language sentence $s^{(2)}_{1:n}$:
\begin{equation*}
    x=\left[s^{(1)}_1,\cdots,s^{(1)}_m,s^{(2)}_1,\cdots,s^{(2)}_n\right]\label{eq:tlm_inp}
\end{equation*}

For a given input, TLM follows~\cite{devlin-etal-2019-bert}, and selects a random set of input tokens $y=\{s^{(l)}_1,\dots,s^{(l)}_k\}$ for masking. Let us denote the masked input sequence with $\tilde x$, and the ground-truth targets for masked positions with $\hat y$.
TLM employs the masked language modelling (MLM) objective to maximise the log-probability of correct labels $\hat y$, conditioned on the masked input $\tilde x$:
\begin{equation*}
\mathcal{L} = {\frac{1}{\abs{\mathcal{X}}}}\sum_{x\in\mathcal{X}}\log\Pr(\hat y{\vert}\tilde x; \theta)\label{eq:loss}
\end{equation*}
where $\theta$ are the model parameters. We keep the standard hyper-parameters for masking, \ie $15\%$ of inputs are randomly selected for masking, from which $80\%$ are replaced with the \mask token, $10\%$ are replaced with random tokens from the vocabulary, and $10\%$ are left intact.

\subsection{Visual translation language modelling}
\label{sec:vtlm}
VTLM extends the TLM by adding the visual modality alongside the translation pairs (Figure~\ref{fig:VTLM}). Therefore, we assume the availability of sentence pair \& image triplets and redefine the input as:
\begin{equation*}
    x\mathrm{=}\left[s^{(1)}_1,\cdots,s^{(1)}_m,s^{(2)}_1,\cdots,s^{(2)}_n,\textcolor{purple}{v_1,\cdots,v_o}\right]\label{eq:vtlm_inp}
\end{equation*}
where $\{v_1,\cdots,v_o\}$ are features extracted from a Faster R-CNN model~\cite{fasterrcnn} pre-trained on the Open Images dataset~\cite{openimages}.\footnote{The ``\I{faster rcnn inception resnet v2 atrous oid v4}'' model from TensorFlow.} Specifically, we extract convolutional feature maps from $o=36$ most confident regions, and average pool each of them to obtain a region-specific feature vector $v_i \in \mathbb{R}^{1536}$. Each region $i$ is also associated with a detection label $\hat v_i$ provided by the extractor. Before encoding, the feature vectors and their bounding box coordinates are projected into the language embedding space.

The final model processes translation pairs and projected region features in a \B{single-stream} fashion~\cite{vlbert,unicodervl}, and combines the TLM loss with the masked region classification (MRC) loss as follows:
\begin{equation*}
\mathcal{L} = {\frac{1}{\abs{\mathcal{X}}}}\sum_{x\in\mathcal{X}}\log\Pr(\{\hat y, \textcolor{purple}{\hat v}\}{\vert}\tilde x; \theta)\label{eq:vtlm_loss}
\end{equation*}

\paragraph{Masking.} $15\%$ random masking ratio is applied separately to both language and visual streams, and the \textcolor{purple}{$\hat v$} above now denotes the correct region labels for the masked feature positions. Different from previous work that zeroes out masked regions~\cite{tan-bansal-2019-lxmert,vlbert}, VTLM replaces their projected feature vectors with the \mask token embedding.\footnote{Although this choice is mostly practical, we hypothesise that using the same signal for both language and visual masking can be beneficial for grounding.} Similar to textual masking, $10\%$ of the random masking amounts to using regional features randomly sampled from all images in the batch, and the remaining $10\%$ of regions are left intact.

\subsection{Pre-training}
VTLM requires a three-way parallel multimodal corpus, which does not exist in large-scale. To address this, we extend\footnote{\url{https://hucvl.github.io/VTLM}} the \B{Conceptual Captions} (CC)~\cite{conceptual} dataset with German translations. CC is a large-scale collection of $\sim$3.3M images retrieved from the Internet, with noisy \I{alt-text} captions in English. The translation of English captions into German was automatically performed using an existing NMT model~\cite{ng-etal-2019-facebook} provided\footnote{The \T{transformer.wmt19.en-de} model.} in the Fairseq~\cite{fairseq} toolkit. Since some of the images are no longer accessible, the final corpus' size is reduced to $\sim$3.1M triplets.
We used byte pair encoding (BPE)~\cite{sennrich-etal-2016-neural} to learn a joint 50k BPE model on the CC dataset. The pre-training was conducted for $1.5$M steps, using a single RTX2080-Ti GPU, and best checkpoints were selected with respect to validation set accuracy.

\begin{table*}[t!]
\centering
\resizebox{.77\textwidth}{!}{%
\begin{tabular}{lcccccc@{}}
\toprule
& \MC{2}{c}{\B{\small 2016}} & \MC{2}{c}{\small \B{2017}} & \MC{2}{c}{\small \B{COCO}} \\
& \MC{1}{c}{\textsc{Meteor}}
& \MC{1}{c}{\textsc{Bleu}}
& \MC{1}{c}{\textsc{Meteor}}
& \MC{1}{c}{\textsc{Bleu}}
& \MC{1}{c}{\textsc{Meteor}}
& \MC{1}{c}{\textsc{Bleu}}
\\ \midrule
\MC{7}{l}{\textcolor{tabtitle}{Best RNN-MMT}~\cite{ozan-tez}}   \\
      &  58.7 &  39.4 &  52.9 &  32.6 &  --  &  --   \\
\MC{7}{l}{\textcolor{tabtitle}{Graph-based Transformers MMT}~\cite{graph_based_att}}   \\
      &  57.6 &  39.8 &  51.9 &  32.2 & 37.6 &  28.7 \\
\MC{7}{l}{\textcolor{tabtitle}{\B{Ensemble} RNN-MMT}~\cite{delbrouck-dupont-2018-umons}}   \\
      &  59.6 &  40.3 &  --   &  --   &  --  &  --   \\
\MC{7}{l}{\textcolor{tabtitle}{\B{Unconstrained} Transformers MMT}~\cite{helcl-etal-2018-cuni}}   \\
      &  59.1 &  42.7 &  --   &  --   &  --  &  --   \\
\midrule
\MC{7}{l}{\textcolor{tabtitle}{Our Baseline Transformers \B{(from scratch)}}}   \\
\T{NMT}       &  56.4 &  37.6 &  51.3 & 30.9  & 47.2 &  27.5 \\
\T{\,\,+CC}   &  58.8 &  39.5 &  55.6 & 36.2  & 51.5 &  33.0 \\
\T{MMT}       &  55.4 &  35.2 &  49.5 & 27.7  & 46.2 &  25.4 \\
\midrule
\MC{7}{l}{\textcolor{tabtitle}{\B{VTLM}: Pre-train and fine-tune on \B{Multi30k}}}   \\
\T{MMT}   & 59.0 & 40.2 & 53.5 & 32.7 & 49.3 & 28.9  \\
\midrule
\MC{7}{l}{\textcolor{tabtitle}{\B{TLM}: Pre-train on \B{CC} -- fine-tune on \B{Multi30k}}}   \\
\T{NMT}   & 60.7             & 43.1             & 56.5            & 37.6             & 53.3             & 34.8              \\
      & \sta{60.5}{0.21} & \sta{42.5}{0.46} & \sta{56.4}{0.10} & \sta{37.3}{0.38} & \sta{53.1}{0.13} & \sta{34.6}{0.17}  \\[.3ex]
\T{MMT}   & 60.3    & 41.9      & 56.7      & 37.6      & 53.3      & 34.3          \\
      & \sta{60.2}{0.08} & \sta{41.7}{0.18} & \sta{56.5}{0.16}& \sta{37.5}{0.10}  & \sta{53.0}{0.20} & \sta{34.1}{0.14}  \\
\midrule
\MC{7}{l}{\textcolor{tabtitle}{\B{VTLM}: Pre-train on \B{CC} -- fine-tune on \B{Multi30k}}}   \\
\T{NMT}   & 61.2     & 43.3     & 56.9     & 37.2     & 53.7     & 35.1  \\
      & \sta{60.5}{0.46} & \sta{42.5}{0.53} & \sta{56.4}{0.34} & \sta{37.0}{0.16} & \sta{53.1}{0.42} & \sta{34.6}{0.40}  \\[.3ex]
\T{MMT}   & 60.8 & 42.7 & 57.1 & \B{38.1} & 53.1 & 34.2  \\
      & \sta{60.6}{0.15} & \sta{42.6}{0.14} & \sta{56.9}{0.20} & \sta{37.7}{0.43} & \sta{53.0}{0.05} & \sta{33.9}{0.19}  \\
\MC{7}{l}{\textcolor{tabtitle}{\B{VTLM}: Alternative (0\% visual masking during pre-training)}}  \\
\T{MMT}    & \B{61.3} & \B{44.0} & \B{57.2} & 38.0 & \B{53.8} & \B{35.2}  \\
      & \sta{60.9}{0.30} & \sta{43.3}{0.59} & \sta{57.1}{0.07} & \sta{37.6}{0.31} & \sta{53.6}{0.17} & \sta{35.1}{0.09}  \\
\bottomrule
\end{tabular}%
}
\caption{Quantitative comparison of experiments: when the mean and the standard deviation is reported, the single numbers appearing above, denote the maximum across three different runs.}
\label{tbl:results}
\end{table*}

\paragraph{Settings.} We use a small version of the TLM~\cite{xlm}\footnote{https://github.com/facebookresearch/XLM} and set the model dimension, feed-forward layer dimension, number of layers and number of attention heads to $d=512$, $f=2048$, $l=6$ and $h=8$, respectively. We randomly initialise model parameters, instead of using pre-trained LM checkpoints such as BERT or XLM. We use Adam~\cite{adam} with the mini-batch size and the learning rate set to $64$ and $0.0001$, respectively. The dropout~\cite{dropout} rate is set to $0.1$ in all layers. The pre-training is done for $1.5$M steps using a single RTX2080-Ti GPU, and best checkpoints are selected with respect to validation accuracy.

\subsection{Baseline MT models and fine-tuning}
Our experimental protocol consists of \B{initialising} the encoder and the decoder of Transformer-based NMT and MMT models with weights from TLM/VTLM, and \B{fine-tuning} them with a smaller learning rate. The architectural difference between the NMT and the MMT models is that the latter encodes $36$ regional visual features as part of the source sequence, similar to the VTLM~($\S$~\ref{sec:vtlm}). As a natural baseline, we train constrained (trained only on the MT dataset) models without transferring weights from the pre-trained TLM/VTLM models. We refer to these models as \B{from-scratch}. For the fine-tuning experiments, we train three runs with different seeds. For evaluation, we use the models with the lowest validation set perplexity to decode translations with beam size equal to 8.

\paragraph{Dataset.} We use the standard MMT corpus \B{Multi30k}~\cite{elliott-etal-2016-multi30k} for both fine-tuning and from-scratch runs. It contains 30k image descriptions from Flickr30k~\cite{flickr30k} and their human translations in German for training, along with three test sets of 1K samples each: the original and the most in-domain \B{2016} test set, as well as \B{2017} and \B{COCO} test sets created using images and descriptions collected from sources other than Flickr.

\paragraph{Settings.} For fine-tuning, we use the same hyper-parameters as the pre-training phase, apart from decreasing the learning rate to $1e\rm{-}5$. For MT models that are trained from scratch, we increase the dropout rate to $0.4$ and linearly warm up the learning rate from $1e\rm{-}7$ to $1e\rm{-}4$ during the first 4,000 iterations. \I{Inverse square-root} annealing is applied after 4,000 iterations.


\vspace*{4mm}
\section{Results}
\label{sec:results}
\subsection{Machine translation}
Table~\ref{tbl:results} reports \textsc{Meteor} and \textsc{Bleu} scores across three different test sets of Multi30k. First, we observe that the MMT system trained from scratch is consistently worse than its NMT counterpart.
However, the gap disappears when pre-trained TLM/VTLM checkpoints are fine-tuned for MT. This suggests that pre-training may be necessary for \I{single-stream} multimodal encoding, where the number of regions ($36$) outnumbers the avg.\  number of source tokens ($13$ for Multi30k).

Second, we see that the best performances are obtained when models are first pre-trained on the \I{three-way} parallel Conceptual Captions (CC) dataset. To validate this further, we train a baseline NMT on the concatenation of Multi30k and CC (NMT+CC) and an MMT that uses only Multi30k for \B{both} pre-training and fine-tuning. The results clearly show that these systems lag behind the ones pre-trained on CC.

We also experimented with an alternative pre-training strategy where we still have the MRC task to predict the object labels of randomly selected visual regions but the input regional features for those positions are not actually replaced with \mask .
In other words, we let the model predict the object labels using a cross-lingual and multimodal input where only input words are randomly masked. Interestingly, this alternative MMT (Table~\ref{tbl:results}) reveals that not masking visual regions during pre-training yields slightly better results overall. 
Overall, MMT fine-tuning on VTLM sets a new state of the art across all Multi30k test sets.\footnote{We exclude~\citet{gronroos-etal-2018-memad} as their improvements (45.5 BLEU) were not due to multi-modality but rather to other modifications such as heavy parallel data augmentation, domain fine-tuning, and ensembling.} We leave the exploration of visual region masking for the MRC task as future work and proceed with the alternative variant in the following experiments.

\paragraph{Encoder attention parameters.}
When fine-tuning the TLM for MT, the default XLM implementation randomly initialises the decoder's missing encoder attention parameters. In our experiments, we noticed that \B{copying} those parameters from the TLM self-attention layers substantially improves the results up to $2.2$ BLEU.

\subsection{Explicit masking}
Here, we will evaluate the extent to which the visual information is taken into account (i) when TLM/VTLM predicts masked tokens, and (ii) when the fine-tuned NMT and MMT models are forced to translate source sentences with missing visual entities. For the latter, we use Flickr30k entities~\cite{Flickr30k_entities} to mask head nouns in 2016 test set sentences, similar to~\citet{caglayan-etal-2019-probing}.


\begin{table}[t!]
\centering
\resizebox{.99\columnwidth}{!}{%
\begin{tabular}{@{}lcccccc@{}}
\toprule
& \MC{3}{c}{\textsc{Valid}} & \MC{3}{c}{\textsc{Test}}      \\
\cmidrule(l){2-4}
\cmidrule(l){5-7}
    & \textsc{En}    & \textsc{De}    & \textsc{Both} 
    & \textsc{En}    & \textsc{De}    & \textsc{Both}  \\
\cmidrule(l){2-4}
\cmidrule(l){5-7}
\T{TLM}   & 89.0      & 87.3      & 55.2      & 88.5      & 86.3      & 53.6      \\
\T{VTLM}  & \ua{0.9}  & \ua{1.4}  & \ua{5.0}  & \ua{1.1}  & \ua{2.2}  & \ua{5.8}  \\
\T{\,\,+shuf}      & \da{1.0}  & \da{0.2}  & \da{7.7}  & \da{1.3}  & \da{0.3}  & \da{7.4}  \\
\bottomrule
\end{tabular}%
}
\caption{Masked \I{last-word} prediction accuracies: VTLM gains are with respect to TLM, whereas the incongruent (+shuf) drops are relative to VTLM.}
\label{tbl:probes}
\end{table}
\paragraph{Last-word masking.} In this experiment, we measure the target word prediction accuracy, when last tokens\footnote{We pre-process the sentences to ensure that they do not end with punctuation marks, which would make the task easier for masked punctuation.} of input caption pairs are systematically masked during evaluation. Table~\ref{tbl:probes} suggests that the visual information is much more helpful (\ie up to 6\% accuracy improvement) when last tokens are masked in both English and German captions. However, if one caption is available, it provides enough context for cross-lingual prediction. Finally, when we shuffle (+shuf) the test set features to introduce incongruence~\cite{elliott-2018-adversarial}, we see that the VTLM model deteriorates substantially. This confirms that the accuracy improvements are not due to side-effects of experimentation noise, such as regularisation or random seed related effects.

\begin{table}[t!]
\centering
\resizebox{.75\columnwidth}{!}{%
\begin{tabular}{rcc}
\toprule
& \textsc{Mask}  & \textsc{Remove} \\ \midrule
\T{TLM\ra NMT}   & 31.44     & 27.38     \\
\T{TLM\ra MMT}   & \da{0.43} & \da{0.26} \\
\midrule
\T{VTLM\ra NMT}  & 31.27     & 27.63     \\
\T{VTLM\ra MMT}  & \ua{1.65} & \ua{0.65} \\
\bottomrule
\end{tabular}%
}
\caption{Entity masking on 2016 test set: results are BLEU averages of three fine-tuned MT systems.}
\label{tbl:res_entity_mask}
\end{table}

\paragraph{Entity masking in MT.} 
We devise two ways of masking entities \ie we either replace them with the \T{[MASK]} token or remove them entirely so that the masking phenomena is \I{not known} to the model. 
The results in Table~\ref{tbl:res_entity_mask} show that MMT models can recover the missing source context to some extent, only when they are pre-trained using the proposed VTLM objective. In other words, the grounding ability can only be acquired when visual modality is present for both pre-training and fine-tuning. The gap between \textsc{Mask} and \textsc{Remove} also seems to highlight the importance of reserving a source position even it is corrupted/masked.

\subsection{Visual attention in MMT}
\begin{figure}[t]
    \centering
    \includegraphics[width=.99\linewidth]{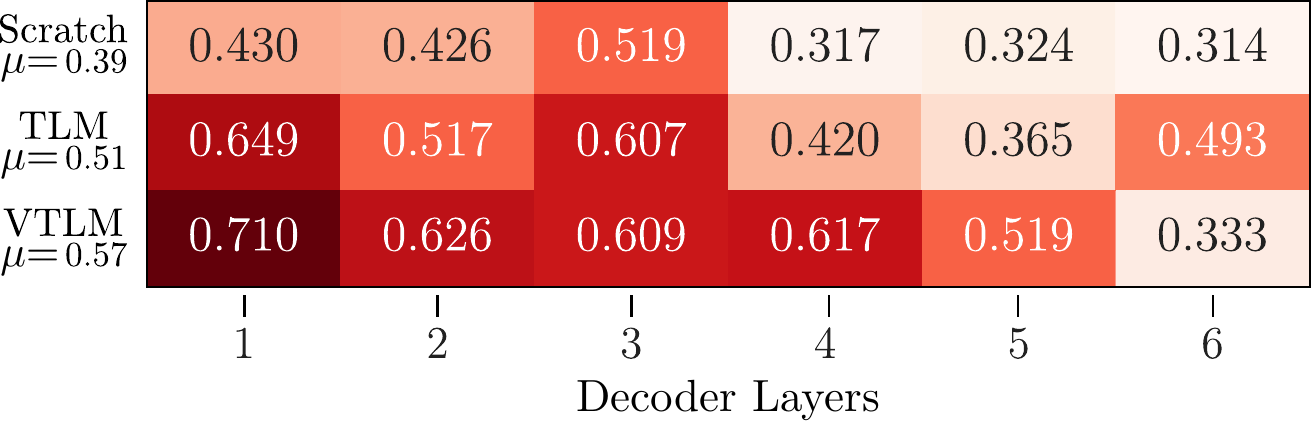}
    \caption{Cross-attention mass over the visual portion of input sequences, averaged across the 2016 test set.}
    \label{fig:mass}
\end{figure}

Here we take the MMT decoder's cross-attention layers and measure the attention mass they attribute to regional features in the input embeddings. Although the encoder's self-attention layers produce increasingly mixed contextual embeddings as we move towards the top layers, \citet{brunner-etal-2002-identifiability} show that the final layer states still encode corresponding input embeddings to some extent. With this assumption at hand, Figure~\ref{fig:mass} shows the average attention mass attributed to the first $36$ (visual) top-layer encoding states, by each cross-attention layer in the decoder. We find these results to be in \I{agreement} with the quantitative metrics (Table~\ref{tbl:results}), with VTLM-MMT assigning substantially more attention to these positions, compared to TLM-MMT and MMT from scratch.

\section{Conclusions}
\label{sec:conclusion}
We proposed a novel cross-lingual visual pre-training approach and tested its efficacy for multimodal machine translation. Our pre-training approach extends the TLM framework~\cite{xlm} with regional features and performs masked language modelling and masked region classification on a three-way parallel corpus. We show that this leads to substantial improvements compared to multimodal machine translation with cross-lingual pre-training only or without pre-training at all. As future work, we consider exploring \I{more informed} masking strategies for visual regions and investigating the impact of visual masking probability for the MRC pre-training task for downstream MMT performance.

\section*{Acknowledgments}

This work was supported in part by TUBA GEBIP fellowship awarded to Erkut Erdem, and the MMVC project funded by TUBITAK and the British Council via the Newton Fund Institutional Links grant programme (grant ID 219E054 and 352343575). Lucia Specia, Pranava Madhyastha and Ozan Caglayan also received support from MultiMT project (H2020 ERC Starting Grant No. 678017) and Lucia Specia from the Air Force Office of Scientific Research (under award number FA8655-20-1-7006).

\bibliography{anthology,eacl2021}
\bibliographystyle{acl_natbib}

\end{document}